\documentclass[conference]{IEEEtran}
\IEEEoverridecommandlockouts
\usepackage{cite}
\usepackage{amsmath,amssymb,amsfonts}
\usepackage{algorithmic}
\usepackage{graphicx}
\usepackage{pgfplots}
\usepackage{stfloats}
\usepackage{subfig}
\usepackage{textcomp}
\usepackage{xcolor}
\def\BibTeX{{\rm B\kern-.05em{\sc i\kern-.025em b}\kern-.08em
    T\kern-.1667em\lower.7ex\hbox{E}\kern-.125emX}}

\usepackage{hyperref}
\usepackage[T1]{fontenc} 
\usepackage[utf8]{inputenc}
\usepackage{verbatim}
\usepackage{float}
\usepackage{multirow}
\usepackage{threeparttable}
\usepackage{footnote}
\usepackage{booktabs}
\usepackage{tikz}

\usepgfplotslibrary{groupplots}
\usetikzlibrary{matrix,positioning}
\pgfplotsset{compat=newest}
\makesavenoteenv{tabular}
\makesavenoteenv{table}

\begin{document}

\title{Summarizing Strategy Card Game AI Competition
\thanks{This work was supported by the National Science Centre, Poland under project number  2021/41/B/ST6/03691.}
}

\author{

    \IEEEauthorblockN{Jakub Kowalski, Rados{\l}aw Miernik}
    \IEEEauthorblockA{
        \textit{Institute of Computer Science, University of Wroc{\l}aw}\\
        Wroc{\l}aw, Poland \\
        Email: \{jakub.kowalski,radoslaw.miernik\}@cs.uni.wroc.pl
    }
    
}

\maketitle
\IEEEpubidadjcol


\begin{abstract}
This paper concludes five years of AI competitions based on Legends of Code and Magic (LOCM), a small Collectible Card Game (CCG), designed with the goal of supporting research and algorithm development. The game was used in a number of events, including Community Contests on the CodinGame platform, and Strategy Card Game AI Competition at the IEEE Congress on Evolutionary Computation and IEEE Conference on Games. LOCM has been used in a number of publications related to areas such as game tree search algorithms, neural networks, evaluation functions, and CCG deckbuilding. We present the rules of the game, the history of organized competitions, and a listing of the participant and their approaches, as well as some general advice on organizing AI competitions for the research community. Although the COG 2022 edition was announced to be the last one, the game remains available and can be played using an online leaderboard arena.
\end{abstract}


\section{Introduction}

Currently, ``grand challenges'' for AI research are not limited to classic boardgames like Chess, Checkers, or Go. 
While they still attract wide attention because of their universal cultural role, it has been shown that modern computer games may serve as milestones for AI development as well.
So far, presented approaches that beat best human players in \emph{Dota~2} \cite{OpenAIDota} and \emph{StarCraft II} \cite{alphastarblog}, are one of the most spectacular and media-impacting demonstrations of AI capabilities. 

The accent is on game features that make designing successful AI players especially difficult, e.g., large action space, long term planning, imperfect information, and randomness.
One game genre containing all these features is Strategy Card Games, also known as Collectible Card Games (CCGs).
Besides the usual AI challenge (successful game-playing),~CCGs have their own like deckbuilding and game-balancing \cite{hoover2020many}.

In recent years, numerous research has been conducted in this domain, assisted by a few related AI competitions. 
The \emph{Hearthstone AI Competition} \cite{dockhorn2019introducing}, with the goal to develop the best agent for the game \emph{Hearthstone} \cite{Blizzard2004Hearthstone} was organized during IEEE Conference in Games in 2018, and the \emph{AAIA’17 Data Mining Challenge: Helping AI to Play Hearthstone} \cite{janusz2017helping} was focused on developing a scoring model for predicting win chances of a player, based on single game state data.

In this paper, we summarize the \emph{Strategy Card Game AI Competition} (SCGAI) organized since 2019 at IEEE Congress on Evolutionary Computation and IEEE Conference on Games. 
The competition is based on \emph{Legends of Code and Magic} (LOCM) \cite{LOCMPage} programming game, designed especially for fair AI vs. AI matches.
LOCM is a small implementation of a CCG, and its advantage over the commercial CCG AI engines is that it is much simpler to handle and thus allows testing more sophisticated algorithms and quickly implementing theoretical ideas.
The Strategy Card Game AI Competition competition aimed to play the same role for the Hearthstone AI Competition as microRTS \cite{Ontanon2013microRTS} plays for various StarCraft AI contests \cite{churchill2016starcraft}. That is, encourage advanced research, free of drawbacks of working with the full-fledged game. 

The last edition of the SCGAI Competition took place in 2022, and the goal of this publication is to summarize it as a whole.
We start with establishing its place in the context of other CCG-related competitions, defining the characteristics behind its uniqueness, and presenting related research based on the LOCM game.
We describe the rules of the game and the course of its development, showing which aspects of the game, and why, were updated during the contest's lifespan.
We present a concise history of the competition, pointing out the characteristics of each edition, and listing the approaches taken by the competitors, in particular, the winners.
Finally, we share our thoughts and experiences related to the competition that might be helpful to AI competition organizers in general.

\section{Related Research}

We shortly present a summary of CCG-related research (not originating in LOCM and SCGAI), focusing on Hearthstone-based ones; a game used for other academic AI competitions. 

\subsection{Hearthstone AI Competition}

The Hearthstone AI Competition \cite{dockhorn2019introducing} was held three times, from 2018 to 2020, at IEEE Conference on Games. 
Each year it received between 30 and 50 submissions, divided between two tracks.
The \emph{Premade Deck Playing} track required using one of the decks prepared by the organizers: 6 decks were known upfront, while an additional 3 were used only for the final evaluation.
The \emph{User Created Deck Playing} track allowed agents to prepare their own deck, where the contestants used popular, user-created decks known to the Hearthstone players.
 
The competition was based on SabberStone -- a Hearthstone simulator written in C\# .Net Core that claimed to implement, as of 2019, when its development stopped, 98\% of the base cards from the game.
Creating an agent required implementing a C\# class with a method that receives a game state and returns an action to perform.
The complexity was reflected in a substantial time budget: 30 seconds per turn.

Winning strategies of the submitted agents were mostly based on search algorithms: Rolling Horizon Evolution \cite{Liu2016Rolling}, MCTS \cite{Browne2012ASurvey}, Pruned BFS, or Dynamic Lookahead algorithm; usually paired with a state evaluation function.
For example, the best agent of 2019 was based on Information Set MCTS and sparse sampling \cite{Choe2019EnhancingMonteCarlo}.
Runner-up in 2018 used competitive coevolutionary optimization for learning heuristic evaluation function used in a greedy one-step look-ahead algorithm \cite{garcia2020optimizing}.

\subsection{Hearthstone Data Mining Challenge}

Another interesting, although a single-time event, was the beforementioned \emph{Data Mining Challenge: Helping AI to Play Hearthstone}, organized at the International Symposium on Advances in Artificial Intelligence and Applications in 2017 \cite{janusz2017helping}. 
It lasted less than two months and attracted 188 submissions.

The dataset provided to participants contained examples of game states extracted from Hearthstone playouts between random AI players.
The goal was to predict the winning probability of the first player based on game states and submit their predictions to the Knowledge Pit competition platform \cite{janusz2015knowledge}.
The training set given to the participants consisted of 3,250,000 game states. The test set used for final evaluation contained 750,000 states, and the results on 5\% of it were known for the submitting contestants as a preliminary score.

All top-ranked classifiers used neural networks.
The winning solution used an ensemble over a few variants of convolutional neural networks \cite{grad2017helping}.
The runner-up solution was based on Logistic Regression combined with extreme gradient-boosted decision trees and deep learning \cite{vu2017ensemble}.


\subsection{CCG Game playing}

Although a variety of approaches were tried, most of them took the form of MCTS \cite{Browne2012ASurvey} enhancements.
The algorithm seems to work well in such a stochastic, multi-action environment, although the size of the games inspires the development of methods for reducing the action space \cite{Choe2019EnhancingMonteCarlo}.
Among many improvements described in \cite{zhang2017improving}, Card-Play Policy Networks can improve rollout quality and reduce their branching factor.

However, full rollouts are too noisy.
Thus, search is usually combined with some form of state evaluation based on expert knowledge and heuristics \cite{santos2017monte} or neural networks \cite{swiechowski2018improving}.

A recent spectacular success in CCG AI is winning against the top 10 human player of the official Hearthstone League in China \cite{Xiao2023MasteringHS}.
Its authors also won the last SCGAI edition; their submission is briefly described in subsection \ref{ssec:byterl}.

\subsection{CCG deckbuilding and game balancing}

These are two important tasks closely related to each other. 
As the goal of deckbuilding is to provide a combination of cards that will be able to consistently win against a variety of opponents, game balancing can be seen as a method of ensuring that the set of such successful decks and strategies will be sufficiently large and diverse.

The usual approach for these tasks is to use some form of evolution, treating cards as genes and decks as genotypes, with evaluation 
based on playouts between AI agents using these decks.
These include 
standard evolutionary algorithms \cite{garcia2016evolutionary},  
Evolutionary Strategies \cite{bhatt2018exploring}, MAP-Elites \cite{Fontaine2019Mapping,zhang2022deep}.
An example of multiobjective EA for game balancing focused on finding overperforming cards can be found in \cite{Silva2019evolving}.


An approach tailored to the arena game mode in LOCM, extending EA with active genes to improve learning efficiency, was described in \cite{Kowalski2020EvolutionaryApproach}.
Another study analyzes the influence of representation and the choice of opponent used to test the model on the quality of learned heuristics \cite{Miernik2022EvolvingEvaluation}.

%

\section{Legends of Code and Magic}

LOCM is a CCG designed for AI research.
In comparison to real-world CCGs, it has only a handful of mechanics, and all card effects are deterministic.
While battling, the only source of non-determinism is one's deck order.

In total, there were three versions of the game used for the competitions: 1.0, 1.2, and 1.5.
Each version changed the game in a backward-incompatible fashion, slightly increasing the complexity.
The detailed rules are described below.

For each version, the organizers provided an online arena available on CodinGame as well as an offline Java referee and two faster implementations of version 1.2 in Nim and Rust.
Additionally, the authors of \cite{vieira2020drafting,Vieira2022Exploring} implemented a set of OpenAI Gym environments.

\subsection{Version 1.0}

\begin{figure}[t]
    \centering
    \includegraphics[width=\columnwidth]{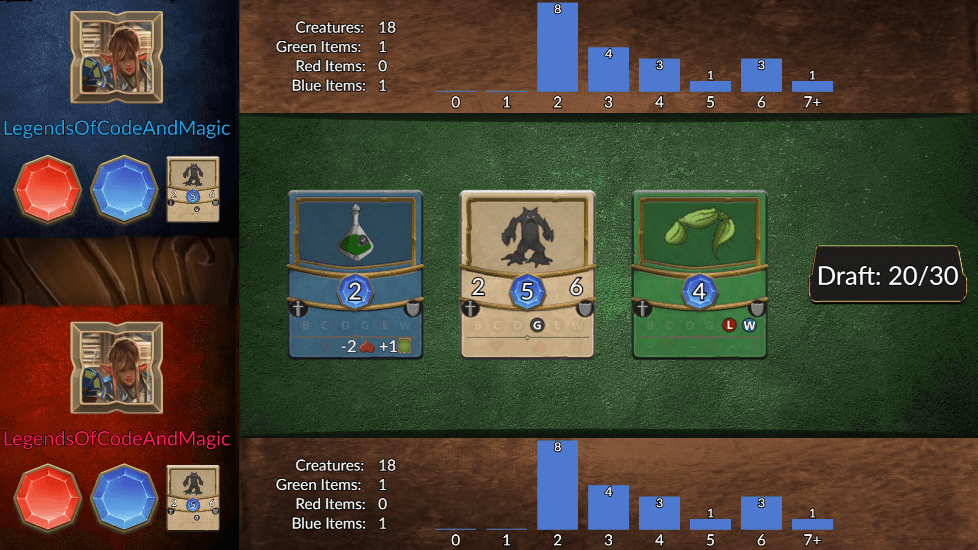}
    \caption{Draft phase in version 1.0 and 1.2 of LOCM. Available cards are in the center. Above and under it are players' decks' statistics and their mana curves.}
    \label{fig:12draft}
\end{figure}

Each match starts with a \textit{draft phase}, where the players build their decks in a fair arena mode.
For 30 turns, they select one out of three cards (both players share the same options).
The UI includes players' decks' statistics and their \textit{mana curve} (histogram of cards' costs), as shown in Figure~\ref{fig:12draft}.

\begin{figure}[t]
    \centering
    \includegraphics[width=\columnwidth]{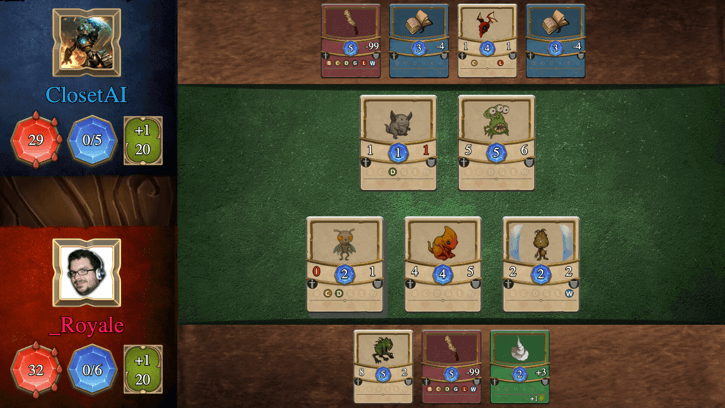}
    \caption{Battle phase in version 1.0 of LOCM. Basic players' info is on the left; their hands and the board are on the right.}
    \label{fig:10battle}
\end{figure}

Next, the \textit{battle phase} begins.
Both players start with one \textit{mana}, used to play the cards.
To account for the first player's advantage, the second player receives one additional mana each turn, as long as they will not use all of it in one turn.
The UI of the battle phase is shown in Figure~\ref{fig:10battle}.

Every player starts with thirty health, five \textit{runes}, corresponding to 25, 20, 15, 10, and 5 health thresholds, respectively.
When the player's health reaches the threshold, the rune breaks and grants an additional card draw for the next turn.

Each turn starts with increasing the max mana by one (up to a maximum of 12), recharging mana, and drawing cards (one plus additional draw), up to a maximum of 8 in \textit{hand}.
If there are no cards to draw, the player loses a rune instead, and their health is reduced to its threshold.
After 50 turns, both decks are considered empty.

Next, the player can play cards if they have enough mana, attack with their creatures, and finally end their turn.
The game ends when at least one player's health drops to zero or below.

Every card is either a \textit{creature} or an \textit{item}.
The former can be \textit{summoned} on the board; the latter \textit{used} on a target.
All cards share three basic attributes (attack, cost, defense), three effects (bonus draw, own health gain, and opponent's health reduction), and a subset of \textit{keywords}.

Keywords are creatures' special abilities and take effect when they battle.
There are six of them: \emph{breakthrough} (deal excess damage to the opponent), \emph{charge} (summoned creature can attack immediately), \emph{drain} (dealt damage heals the owner), \emph{guard} (must be attacked first), \emph{lethal} (kills damaged creature instantly), and \emph{ward} (blocks all incoming damage once).

Creature cards stay on the board as long as their defense is positive.
Starting from the turn after they were summoned, they can attack the opponent or their creatures once each turn.
While battling, both creatures attack simultaneously.

Item cards are subdivided into three colors: green, red, and blue.
Green items can be used on own creatures, increasing their statistics and adding new keywords.
Red items can be used on enemy creatures, reducing their statistics and removing keywords.
Blue items are similar to red, but additionally can be used directly on the opponent.

\subsection{Version 1.2}

\begin{figure}[t]
    \centering
    \includegraphics[width=\columnwidth]{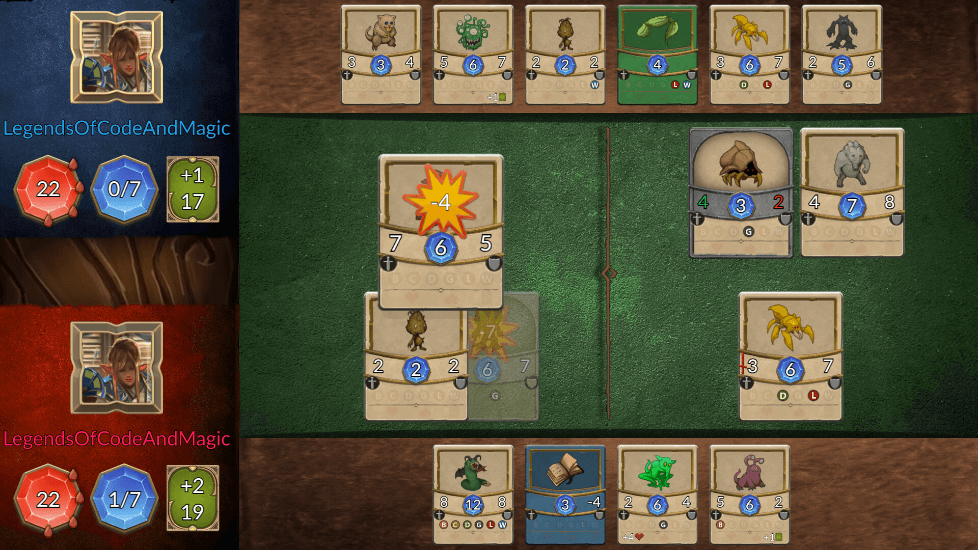}
    \caption{Battle phase in version 1.2 and 1.5 of LOCM. Contrary to version 1.0, the board is now split into two lanes.}
    \label{fig:12battle}
\end{figure}

In contrary to Hearthstone, The Elder Scrolls: Legends have two lanes, i.e., the board is split into two parts.
And while LOCM is based on the latter, version 1.0 had a single board of size 6 to simplify the game.
Version 1.2 changes that, splitting the board into two lanes of size 3, as shown in Figure~\ref{fig:12battle}.

This not only changes the size and shape of the game tree but also impacts the importance of certain keywords.
For example, creatures with \textit{Guard} now protect only half of the board, and the ones with \textit{Lethal} have fewer targets.

\subsection{Version 1.5}

The main conclusion from versions 1.0 and 1.2 is that an agent can achieve amazing results while hardcoding the entire draft phase.
This alone degenerated the game to a single phase.

Replacing the draft with a construction phase alone would not solve this issue, as knowing all the available cards, the agents could tackle the deck construction problem offline.
That is what happened in the Hearthstone AI Competition.

To overcome this problem, version 1.5 generates a new card set for each match.
It forces agents to generalize their play style to all possible cards, including unbalanced ones.
Some can effectively win most matches instantly when played (e.g., a blue item dealing 99 damage with zero cost); others can be useless (e.g., a red item dealing no damage with no keywords).
Both can be used to test agents' deck construction capabilities.

\begin{figure}[t]
    \centering
    \includegraphics[width=\columnwidth]{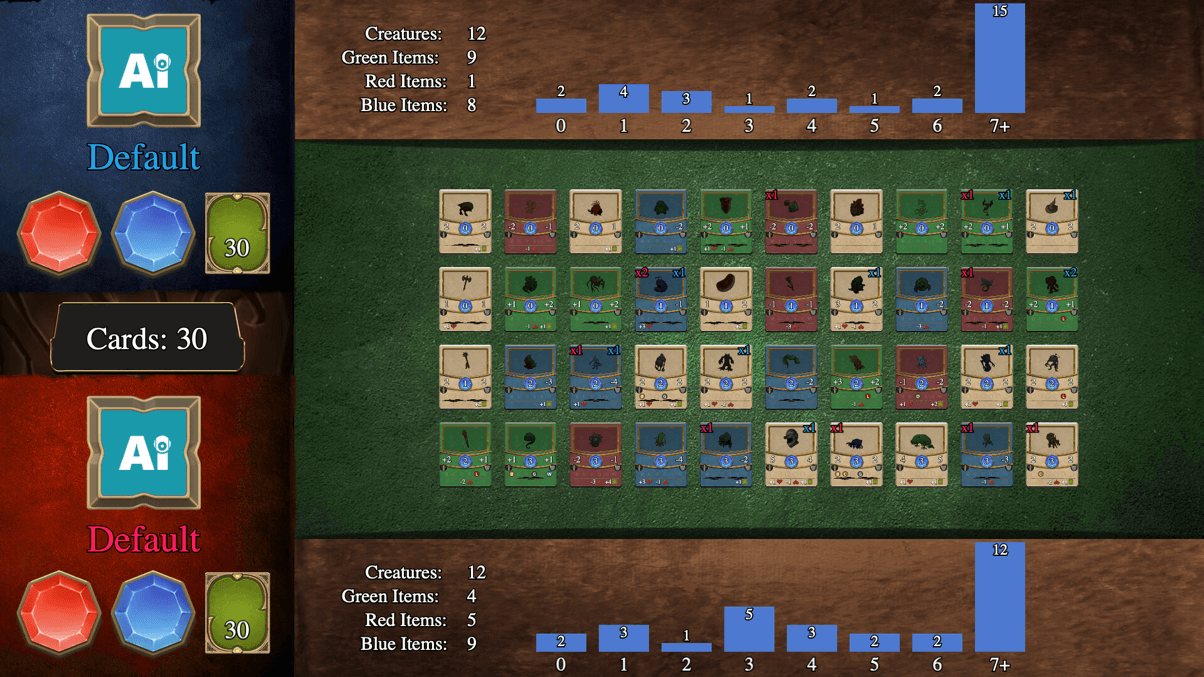}
    \caption{Construction phase in version 1.5 of LOCM. The red and blue numbers in the card's top left and right corners indicate how many copies of it each player picked.}
    \label{fig:15construction}
\end{figure}

Technically, the construction phase is a single, four second long turn, where the agents are presented with 120 cards.
They can pick up to 30 cards, using at most two copies of each.
In the UI, cards are shown in three frames, as shown in Figure~\ref{fig:15construction}.

Version 1.5 also introduced a new \textit{Area} ability.
For creature cards, it either added an extra copy on the same lane (\textit{Lane1} value) or the other lane (\textit{Lane2} value), if there is space for it.
For item cards, it either applied it to all creatures in the target's lane and side of the board (\textit{Lane1}), or all creatures on the target's side of the board (\textit{Lane2}).
The default value (\textit{Target}) has no special behavior.

Finally, version 1.5 removed the runes mechanic completely, as it was unreasonably complex for how it worked.
In return, players get to draw an additional card for every 5 health lost in the previous round, preserving the aid while losing health.

\section{Competition}

\subsection{Legends of Code and Magic on CodinGame (2018), v1.0}


CodinGame is a challenge-based coding platform offering (among others) tens of multiplayer bot programming games.
More than 25 programming languages, communication based on standard input/output using game-specific text protocols, and an in-browser coding environment allowed it to gather a sizable agent programming community.

The first LoCM competition, based on version 1.0, was only 24 hours long (\textit{Sprint}) and gathered 742 participants\footnote{\href{https://www.codingame.com/contests/legends-of-code-and-magic/leaderboard}{CodinGame Sprint, July 25, 2018}}.
The second one started two days later, lasted 30 days (\textit{Marathon}), and received 2174 submissions\footnote{\href{https://www.codingame.com/leaderboards/challenge/legends-of-code-and-magic-marathon/global}{CodinGame Marathon, July 27, 2018}}.

The platform's moderators, as well as the community itself, highly discourage open sourcing full agents' code, and these are not generally available.
However, players share their strategies and detailed information about their thought process in so-called \textit{post-mortems} on the platform's forum\footnote{\href{https://www.codingame.com/forum/t/legends-of-code-magic-cc05-feedback-strategies/50996}{Legends of Code and Magic Feedback \& Strategies}}.



The draft phase was dominated by handcrafted or experimentally adjusted heuristics that can be effectively implemented as a fixed ordering of cards.
As it is possible to compete with everyone's agent using a given game seed, many players mimicked the top players' ordering.
Some players explored applying a \textit{mana curve} (balancing the number of cards with a given cost) but with no significant benefits.

Handcrafted rule-based agents dominated the battle phase.
However, the best players employed variants of well-known search methods like Minimax (few plys deep; alpha-beta and heuristic pruning) and MCTS (depth-limited with a heuristic cut-off).
The most significant improvements reported by all players were move ordering and pruning, and lethal (winning) move detection.

\subsection{Strategy Card Game AI Competition (2019-2021), v1.2}

The competition was no longer ran on the CodinGame platform, and thus the agent limitations changed.
Most importantly, there are no programming language restrictions -- the only requirement is compatibility with a UNIX-based system.
The memory limit got lowered (256MB; agents using 1024MB or more were disqualified), and the time limit for the standard battle turns got doubled.

To reduce the noisiness of the results, all players played using a fixed number of randomly sampled decks ten times on the same random seed, resulting in identical card ordering.
In such a setting, two deterministic agents would achieve the same results in all ten games.
Additionally, every match was mirrored to account for the difference between being the first or second player.

\subsubsection{CEC 2019}

The first competition received six submissions, and all of them were notably stronger than the baselines provided by the organizers.
Four were rule-based agents with a variety of heuristics; two performed a proper search.
The winner, \textit{Coac}, based its battle phase on a minimax-like search of depth three with alpha-beta and heuristic pruning, turned out to be significantly stronger than all of the competitors (33\% higher win rate than the runner-up).
The draft phase used a fixed ordering of cards (the highest card was selected).

\subsubsection{COG 2019}

This competition received three new submissions.
The best one of them, \textit{ProphetCoac}, was an attempt to improve the previous competition's winner by predicting the opponent's hand based on the cards seen during the draft to reduce the branching factor.
Ultimately it did not improve but actually reduced the overall win rate, most likely due to less time available for search.

\subsubsection{CEC 2020}

This competition received one update and two new submissions.
The former was \textit{Coac}, changing the card orderings solely, effectively improving the heuristics.

The new \textit{ReinforcedGreediness} agent was the first to include a neural network.
For the draft phase, it used two networks, one for each side, trained by self-play reinforcement learning.
For the battle phase, it used a best-first search with a heuristic using Bayesian-optimized handcrafted features.

\subsubsection{COG 2020}

This competition received only one new submission.
In this edition, the last year's agents were not evaluated, resulting in a visible change in win rates -- all were lower.
The top two agents from the previous competition switched places, and an in-depth analysis suggests that all agents beat the baselines almost every time while staying relatively competitive (i.e., all agents were decent).

\subsubsection{COG 2021}

This competition received four new submissions.
One of them, \textit{DrainPower}, had two variants -- the default and the aggressive one.
Both shared the same static card ordering for the draft and a flat simulation-based algorithm but had different heuristic parameters.
This agent took the first two places, with the aggressive version being slightly better.

\subsection{Strategy Card Game AI Competition (2022), v1.5}

As the draft phase was replaced with a deck construction one and a new area effect was introduced, the game protocol did change, and thus all of the previously submitted agents were no longer compatible.
Similarly, fixed card orderings are no longer usable, as the cards are now randomly generated for each match.

To let agents perform an in-depth analysis of the cards, the first turn is four seconds long; this is the same as the whole draft phase before.
Other limits were not changed.

The organizers provided only a Java-based referee, updated the CodinGame environment, and the community updated the existing OpenAI Gym environments.

\subsubsection{COG 2022}

This competition was dominated by neural network-based agents -- four of six submissions had one.
Two of them were trained using Proximal Policy Optimization and the other two using other reinforcement learning algorithms.

\renewcommand{\arraystretch}{0.82}
\begin{table}
  \caption{Results of all Strategy Card Game AI competitions. New and updated agents' names are in bold. Baseline agents' names are in italics.}
  \centering
  \begin{threeparttable}
    \begin{tabular}{crrl}
      Year & Place & Win rate & Agent \\
      \toprule
      \toprule
      \multirow{8}{*}{\shortstack{IEEE CEC 2019\\v1.2}}
      & 1 & 94.22\% & \textbf{Coac} \\
      & 2 & 58.93\% & \textbf{UJIAgent2} \\
      & 3 & 50.64\% & \textbf{AntiSquid} \\
      & 4 & 46.72\% & \textbf{Marasbot} \\
      & 5 & 44.69\% & \textbf{UJIAgent1} \\
      & 6 & 42.21\% & \textbf{Conrisc} \\
      & 7 & 37.04\% & \textit{Baseline2} \\
      & 8 & 25.51\% & \textit{Baseline1} \\
      \midrule
      \multirow{11}{*}{\shortstack{IEEE COG 2019\\v1.2}}
      & 1   & 89.88\% & Coac \\
      & 2   & 87.84\% & \textbf{ProphetCoac} \\
      & 3   & 59.50\% & Marasbot \\
      & 4   & 54.44\% & UJIAgent2 \\
      & 5   & 45.04\% & AntiSquid \\
      & 6   & 42.27\% & \textbf{Fabbiamo} \\
      & 7   & 41.12\% & \textbf{UJIAgent3} \\
      & 8   & 39.80\% & UJIAgent1 \\
      & 9   & 37.47\% & Conrisc \\
      & 10  & 32.24\% & \textit{Baseline2} \\
      & 11  & 20.35\% & \textit{Baseline1} \\
      \midrule
      \multirow{13}{*}{\shortstack{IEEE CEC 2020\\v1.2}}
      & 1  & 86.07\% & \textbf{Coac} \\
      & 2  & 79.10\% & \textbf{Chad} \\
      & 3  & 60.20\% & \textbf{ReinforcedGreediness} \\
      & 4  & 58.62\% & ProphetCoac \\
      & 5  & 55.79\% & Marasbot \\
      & 6  & 53.70\% & UJIAgent2 \\
      & 7  & 45.44\% & UJIAgent3 \\
      & 8  & 43.42\% & AntiSquid \\
      & 9  & 42.82\% & Fabbiamo \\
      & 10 & 41.04\% & UJIAgent1 \\
      & 11 & 34.61\% & Conrisc \\
      & 12 & 28.20\% & \textit{Baseline2} \\
      & 13 & 21.00\% & \textit{Baseline1} \\
      \midrule
      \multirow{6}{*}{\shortstack{IEEE COG 2020\\v1.2}}
      & 1 & 79.99\% & Chad \\
      & 2 & 74.68\% & Coac \\
      & 3 & 59.49\% & \textbf{OneLaneIsEnough} \\
      & 4 & 56.21\% & ReinforcedGreediness \\
      & 5 & 16.70\% & \textit{Baseline2} \\
      & 6 & 12.94\% & \textit{Baseline1} \\
      \midrule
      \multirow{11}{*}{\shortstack{IEEE COG 2021\\v1.2}}
      & 1  & 78.72\% & \textbf{DrainPowerAggressive} \\
      & 2  & 77.96\% & \textbf{DrainPower} \\
      & 3  & 75.51\% & Chad \\
      & 4  & 71.77\% & Coac \\
      & 5  & 59.01\% & OneLaneIsEnough \\
      & 6  & 55.38\% & ReinforcedGreediness \\
      & 7  & 37.53\% & \textbf{LANE\_1\_0} \\
      & 8  & 27.17\% & \textbf{Ag2O} \\
      & 9  & 23.56\% & \textit{Baseline1} \\
      & 10 & 21.83\% & \textit{Baseline2} \\
      & 11 & 21.62\% & \textbf{AdvancedAvocadoAgent} \\
      \midrule
      \multirow{7}{*}{\shortstack{IEEE COG 2022\\v1.5}}
      & 1 & 84.41\% & \textbf{ByteRL} \\
      & 2 & 75.00\% & \textbf{NeteaseOPD} \\
      & 3 & 67.57\% & \textbf{Inspirai} \\
      & 4 & 42.04\% & \textbf{MugenSlayerAttack}\tnote{1} \\
      & 5 & 41.09\% & \textbf{USTC-gogogo}\tnote{2} \\
      & 6 & 38.60\% & \textbf{Zylo} \\
      & 7 &  1.29\% & \textit{RandomWItems2lanes} \\
      \bottomrule
    \end{tabular}
    \begin{tablenotes}
      \item [1] MugenSlayerAttackOnDuraraBallV3
      \item [2] Variant ``Zero\_control''
    \end{tablenotes}
  \end{threeparttable}
\end{table}

\section{Players}

There were 22 unique agents submitted in total.
Based on the main algorithm used for playing, we divided them into three groups -- search-based, neural network-based, and other.
Interestingly, most agents in every group have a similar performance characteristic, e.g., they use a similar amount of time and memory while playing.
Three baseline agents provided by the organizers are described in a dedicated subsection.

\subsection{Baselines}

The provided baseline agents were built with two goals in mind: presenting the game rules and providing a better training opponent than a fully random one.
They are fairly trivial and rely on the game engine ignoring invalid actions.

\subsubsection{Baseline 1}
This agent, written in Python, was used for the competitions running version 1.2 of the game.
During the draft phase, it focuses on selecting creature cards with \textit{Guard} keyword and falls back to the first card otherwise.
For the battle phase, it follows a rule-based algorithm focused on using all cards in hand and on the board, attacking the opponent's first and their creatures next.

\subsubsection{Baseline 2}
This agent, written in Python, was used for the competitions running version 1.2 of the game.
During the draft phase, it focuses on selecting creature cards with the highest attack and falls back to the first card otherwise.
For the battle phase, it uses a one-pass algorithm, attacking the opponent using all summoned creatures and summoning all creatures from hand.

\subsubsection{RandomWItems2lanes}
This agent, written in Java, was used for the competitions running version 1.5 of the game.
In the construction phase, it selects all cards at random.
For the battle phase, it uses all green items on its own creatures, attacks the opponent (only creatures with the \textit{Guard} keyword, if there is any), summons all creatures, and finally uses all items.
All actions are targeted at random.

\subsection{Search-based}

Search-based agents were the most common and won all competitions running version 1.2.
All of them employed move pruning either explicitly (certain actions were disallowed) or implicitly (moves were generated out of ordered actions).
Similarly, most of them implemented lethal move detection.

Only some of the agents searched through the opponent's turn, and it is clear that while it is beneficial to do so, it requires further pruning and heuristic evaluation to make it feasible with the exploding branching factor.

\subsubsection{AdvancedAvocadoAgent}
This agent, written in Java, was submitted for COG 2021.
It uses a fixed card ordering for the draft phase and a best-first search for battling.
Weights of the heuristic evaluation function were found offline, using an MCTS-based search over the space of parameters.

\subsubsection{Chad}
This agent, written in Rust, was submitted for CEC 2020 and described in \cite{Witkowski20LOCM}.
It scores the cards using weights computed with harmony search.
The battle phase uses MCTS with the opponent's hand prediction.

\subsubsection{Coac}
This agent, written in C++, was submitted for CEC 2019 and got updated the next year.
For the draft phase, it uses a fixed card ordering.
For the battle phase, a Minimax-like search of depth three (or less, if the time ran out or the tree was too wide) with alpha-beta and heuristic pruning.

\subsubsection{DrainPower}
This agent, written in C\#, was submitted for COG 2021.
Both phases base on heuristic card evaluation.
The battle phase uses a flat simulation-based algorithm, simulating own turn and the opponent's response.
It has two variants -- default and aggressive -- with different weights for the heuristic evaluation function.

\subsubsection{Fabbiamo}
This agent, written in C++, was submitted for COG 2019.
During the draft, it follows a so-called \emph{mana curve}, i.e., it tries to maintain a reasonable number of cards of the same cost.
The battle phase uses a Minimax search of depth four over own actions and depth one of the opponent's.

\subsubsection{LANE\_1\_0}
This agent, written in Java, was submitted for COG 2021.
The draft phase follows a fixed card ordering.
The battle phase uses a flat simulation-based algorithm, simulating only its own turn.

\subsubsection{Marasbot}
This agent, written in C++, was submitted for CEC 2019.
It uses a heuristic evaluation function for both phases and random sampling for the battle phase.
It completely ignores blue items.

\subsubsection{OneLaneIsEnough}
This agent, written in C++, was submitted for COG 2020.
It uses a heuristic evaluation function for both phases and a one-turn deep search, including the opponent's response, for the battle phase.

\subsubsection{Prophet Coac}
This agent, written in C++, was submitted for COG 2019.
It is a modification of the \textit{Coac} agent, including a tentative prediction of the opponent's hand, based on the cards seen during the draft phase and refined by the already played cards.

\subsubsection{UJIAgent3}
This agent, written in Python, was submitted for COG 2019 and partially described in \cite{montoliu2020efficient}.
It uses a fixed card ordering for the draft phase.
For the battle phase, it uses Online Evolutionary Planning \cite{justesen2016online} where the genome encodes a series of actions, and mutation only reorders them.

\subsubsection{Zylo}
This agent, written in Java, was submitted for COG 2022.
It uses a heuristic evaluation function for both phases and a best-first search for the battle phase.
Its parameters were tuned using an evolutionary algorithm.

\subsection{Neural networks-based}

Neural networks-based agents were less common in version 1.2 competitions but dominated the 1.5 one.
While there was no GPU available while playing, some were trained using one.

\subsubsection{Ag2O}
This agent, written in Python, was submitted for COG 2021.
The draft phase bases on card weights trained with q-learning and takes card combinations into consideration.
The battle phase uses a best-first search guided by actions' q-value.
The network composes of four dense layers using $tanh$ as the activation function between and at the end.

\subsubsection{ByteRL}\label{ssec:byterl}
This agent, written in Python, was submitted for COG 2022 and described in \cite{Xi2023MasteringSC}.
Interestingly, it uses only one, end-to-end policy, trained using deep reinforcement learning combined with optimistic smooth fictitious play.
The network architecture is rather complex and includes a Long Short-Term Memory (LSTM) block.

The authors sent an additional version of this agent, adjusted and trained for the 1.2 version of LoCM.
It was compared against all COG 2021 submissions and won by a large margin (more than 20\% higher win rate than the runner-up).

\subsubsection{Inspirai}
This agent, written in Python, was submitted for COG 2022.
Its construction phase uses a heuristic evaluation function adapted from \textit{Coac} and optimized using Bayesian optimization.
The battle phase uses a neural network trained using Proximal Policy Optimization.

The network consists of a dense layer with ReLU activation function, followed by another dense layer.
On top of that, there are two heads using attention \cite{Vaswani2017AttentionIs} for card and target selection.

\subsubsection{NeteaseOPD}
This agent, written in Python, was submitted for COG 2022.
It uses two independent networks trained using Proximal Policy Optimization, one for each phase.

\subsubsection{ReinforcedGreediness}
This agent, written in Python, was submitted for COG 2021.
The draft phase uses a neural network learned by self-play reinforcement learning, trained independently for both sides.
The network consists only of dense layers with a few different activation layers.

The battle phase uses best-first search limited to own turn; the heuristic evaluation function is a linear combination of hand-made features optimized using Bayesian optimization.

\subsubsection{USTC gogogo}
This agent, written in Python, was submitted for COG 2022.
There were two versions -- one using hyperparameters to control both construction and battle phases and the other using model trained using reinforcement learning.
Only the latter was used for the competition.

\subsection{Other}

Other agents were the simplest, usually with a decently sized set of handcrafted rules and heuristic evaluation functions at their core.
Due to their simplicity, most of them acted instantly, within a few milliseconds per turn.

\subsubsection{AntiSquid}
This agent, written in Python, was submitted for CEC 2019.
During the draft, it selects cards using a fixed card ordering, follows a mana curve, and prefers different cards based on the already selected ones.
For the battle phase, it follows a rule-based algorithm, searching for the highest scored sequences of moves.

\subsubsection{Conrisc}
This agent, written in JavaScript, was submitted for CEC 2019.
For the draft, it follows a heuristic evaluation function.
The battle phase uses a rule-based algorithm, using all cards in a given order.
It ignores all items.

\subsubsection{MugenSlayerAttackOnDuraraBallV3}
This agent, written in Python, was submitted for COG 2022.
During the construction phase, it focuses on the cheapest cards with the newly added area effect.
For the battle phase, it follows a list of predefined rules.

\subsubsection{UJIAgent1}
This agent, written in Python, was submitted for CEC 2019.
During the draft phase, it tries to gather a predefined set of cards based on their type and cost.
For the battle phase, it follows a list of predefined rules ordered using a heuristic evaluation function.

\subsubsection{UJIAgent2}
This agent, written in Python, was submitted for CEC 2019.
During the draft phase, it probabilistically tries to gather a predefined set of cards.
For the battle phase, it samples 44 
strategies and picks the one with the highest score.

\section{Takeaways for Competition Organizers}

While an AI competition is a challenge for the contestants, it is also a challenge for the organizers to make it a successful one \cite{Togelius2016HowToRun}. 
It is unclear when a competition becomes successful, but there are several things the organizers have to account for, ranging from coming up with an interesting problem itself to running the final evaluation.

\subsection{Games as test beds}

Modeling problems using games makes them more approachable for most people, especially when the game itself exists (e.g., Hearthstone for the Hearthstone AI Competition) or is a simplified version of one (e.g., LOCM for The Elder Scroll: Legends, MicroRTS for StarCraft).
Additionally, the game's player base is often an excellent source of battle-tested playing strategies and their analyses.

As different games raise different problems, it is crucial to maintain some variety.
A new competition in a game genre not seen before may bring relatively a lot of novel algorithms or methods.
Similarly, extending an existing game can revive stagnated research or make a more versatile benchmark.

At the same time, new problems attract fewer contestants, as they are usually less prestigious or not marketed enough.
Luckily, games are usually flexible, and it is often possible to extend them in a backward-compatible way.
That allows the competition organizers to reuse the previous submissions.

\subsection{Bootstrapping with CodinGame}

The difference in the number of submissions between the CodinGame and academia SCGAI competitions is almost tenfold, while the game rules remained fairly similar across all three versions.
The CodinGame platform has a large and vivid community; using it as an ``incubator'' of an academic competition may be a great way of validating one's idea.

It is important to emphasize that the CodinGame community focuses on solving the problem (i.e., playing the game).
It does not necessarily imply novel algorithms or sophisticated methods.
Actually, the vast majority is the complete opposite: highly optimized versions of well-known algorithms tuned for a specific game or puzzle.

However, CodinGame has some limitations too -- the game engine has to be written in Java, the communication has to be text-based, and the agents are evaluated in a highly restricted environment (1 CPU core, 768MB of RAM, no GPU).
On top of that, the entire game can use at most 30 seconds of summarized agents execution time.

\subsection{Pushing for research}

Arguably, all AI competitions should require submissions to be documented.
On the one hand, it helps the organizers to compare and reason about them, without analyzing the source code.
On the other, it allows future contestants to learn from and improve them, instead of starting from scratch.
Ideally, all agents would result in a paper.

In the first two years of the SCGAI competition, there were two competitions each year.
While it increases the visibility and potentially brings more contestants in, it may result in the opposite -- instead of one competition with six submissions, there will be two with three each.
The SCGAI competition partially solved this issue by automatically resubmitting past agents for future competitions.

Marketing the competition among students may bring many valuable submissions, usually well-documented ones.
Some of them will find participating more interesting than working on an unrelated project, and thus become more involved.
Such projects can evolve into diploma theses and then papers.


Additionally, contestants can be encouraged with prizes.
As the organizers may not want to sponsor them themselves, they can reach out to conference organizers.
IEEE CIS sponsored SCGAI competitions at both CEC and COG conferences.

\subsection{Taming the randomness}


Most real-life games, and so their toy-scale relatives, are highly random.
To ensure fair results, one can use the same game seeds for all agent pairs (e.g., in LOCM, it results in the same cards available during their games).
Additionally, as most agents are non-deterministic on their own, each game seed can be used multiple times to average the result.

Similarly, most games are asymmetrical, giving one of the players a visible advantage.
To account for that, one can run matches in all player configurations using the same game seed.

If a competition runs on multiple machines, or even on one but in parallel, consider interleaving instead of concatenating the results from across runs.
Different agents can utilize the CPU differently, leading to varying results depending on the programs running in the background.

\subsection{Hardware and software}

Comparing all agents pair-wise using a reasonable number of matches requires a notable amount of CPU time.
Luckily, most cloud providers (e.g., DigitalOcean) are keen to sponsor research and related competitions at a small cost of mentioning them or their services while presenting the results.

To let others reproduce the competition results, all of the code used to run it should be published once it concludes.
Most importantly, all kinds of configurations and the dependencies required by all agents should be well documented.

The same applies to the final results.
Win rates and charts are enough for a presentation, but organizers should be fully transparent and share both the raw data and the scripts used to aggregate it.
As the intermediate results can be significant in size, consider compressing them.

Automating the above points allows the organizers to focus on the competition and the AI challenge itself instead of building the infrastructure and tooling for every single contest.
The SCGAI repository meets all these criteria, presenting all agents, tooling, and results in one place.

\section{Conclusion}\label{sec:conclusion}

This paper summarizes five years of Strategy Card Game AI Competition -- an AI programming challenge based on Legends of Code and Magic, a small implementation of a Collectible Card Game authored by Jakub Kowalski and Radosław Miernik.
Its novel fair arena mode stays in opposition to more common collection-based deckbuilding and the simplicity of rules allows search-based approaches to be more profound.
LOCM was first used in CodinGame 5th Community Contest in 2018 and last at IEEE Conference on Games 2022.

To prevent stagnation and introduce rising levels of challenge, the game rules have been extended a few times during this time.
The academia-based editions of the contest gathered 22 challengers, coming with different approaches, from simple rule-based solutions to search-based agents and ones applying deep reinforcement learning. 
The game has been a base for a number of publications concerning playing algorithms and the deckbuilding problem. 

We hope that this summary will serve as a reference point for the competition and related achievements, as well as a general source of knowledge about successful approaches and types of challenges characteristic of the domain of CCGs. 
We also hope that our observations and conclusions will be helpful to other researchers (or companies) that plan to bring to the AI community some new game-oriented challenges.

We especially look forward to the next competitions related to (Collectible) Card Games, as the domain is so broad that in all these years and competitions, it has been studied only superficially, and there are still many challenges left \cite{hoover2020many}. 
One new CCG-related contest we know about is Tales of Tribute AI Competition\cite{Kowalski2023IntroducingTales}, based on the deckbuilding card game Tales of Tribute and advertised for IEEE COG 2023.

Currently, there are no plans to organize the SCGAI competition any further, so it is officially considered close.
However, thanks to CodinGame, all LOCM versions are available online as bot programming games, and everyone can compete against the agents available on the public leaderboards.




\bibliographystyle{IEEEtran} 
\bibliography{bibliography} 

\end{document}